\documentclass[letterpaper, 10 pt, conference]{ieeeconf}  

\IEEEoverridecommandlockouts                              
\usepackage{easyReview}  

\usepackage{graphics} 
\usepackage{epsfig} 
\usepackage{mathptmx} 
\usepackage{times} 
\usepackage{amsmath} 
\usepackage{amssymb}  
\usepackage{mathtools}
\usepackage{booktabs}
\usepackage{threeparttable}
\usepackage{cite}
\usepackage[hidelinks]{hyperref}
\usepackage{algorithm}
\usepackage{algpseudocode}
\usepackage{graphicx}
\usepackage{epsfig}
\usepackage{multirow,makecell}
\usepackage{amsmath}
\usepackage{xcolor}
\usepackage{program}
\let\labelindent\relax
\usepackage{enumitem}
\usepackage{makecell}
\usepackage{fourier} 
\usepackage{array}
\usepackage{units}
\usepackage{tabularx, blkarray}
\usepackage[T1]{fontenc}
\usepackage{hhline}

\begin{document}
\thispagestyle{empty}
\pagestyle{empty}

\begin{titlepage}

\begin{center}
{\huge{IEEE Copyright Notice}}
\end{center}

\bigskip \bigskip

© 2022 IEEE. Personal use of this material is permitted. Permission from IEEE must be obtained for all
other uses, in any current or future media, including reprinting/republishing this material for advertising or promotional purposes, creating new collective works, for resale or redistribution to servers or lists, or reuse of any copyrighted component of this work in other works. 

\bigskip

\begin{center}
{This paper has been accepted for publication in \textit{IEEE Robotics And Automation Letters}~(RA-L).}
\end{center}

\bigskip

\begin{center}
{DOI: \href{https://ieeexplore.ieee.org/document/9714001}{10.1109/LRA.2022.3151396}}
\end{center}

\begin{center}
{IEEE Explore: \url{https://ieeexplore.ieee.org/document/9714001}}
\end{center}

\bigskip

\end{titlepage}

\title{\LARGE{\bf{Concurrent Training of a Control Policy and a State Estimator for Dynamic and Robust Legged Locomotion}}}

\author{Gwanghyeon Ji, Juhyeok Mun, Hyeongjun Kim, and Jemin Hwangbo$^*$
\thanks{This work was supported by Samsung Research Funding \& Incubation Center for Future Technology at Samsung Electronics under Project Number SRFC-IT2002-02.}
\thanks{The Mini Cheetah robot was provided by MIT Biomimetic Robotics Lab and Naver Labs Corporation.}
\thanks{$^*$All authors are with Robotics and Artificial Intelligence Lab in the department of Mechanical Engineering, KAIST, Daejeon 34141, Republic of Korea. {\tt\small jhwangbo@kaist.ac.kr}}
}

\maketitle

\begin{abstract}

In this paper, we propose a locomotion training framework where a control policy and a state estimator are trained concurrently. The framework consists of a policy network which outputs the desired joint positions and a state estimation network which outputs estimates of the robot's states such as the base linear velocity, foot height, and contact probability. We exploit a fast simulation environment to train the networks and the trained networks are transferred to the real robot. The trained policy and state estimator are capable of traversing diverse terrains such as a hill, slippery plate, and bumpy road. We also demonstrate that the learned policy can run at up to \unit[3.75]{m/s} on normal flat ground and \unit[3.54]{m/s} on a slippery plate with the coefficient of friction of 0.22.

\end{abstract}

\section{INTRODUCTION}

In recent years, reinforcement learning~(RL) has become one of the most popular control approaches for legged robots. For quadrupedal robots, there have been remarkable improvements in learning dynamic locomotion skills. Hwangbo et al.~\cite{Hwangbo_2019_LearningAgile} trained control policies for the ANYmal robot~\cite{Hutter_2016_ANYmal} for robust and high-speed locomotion while keeping the balance under large disturbances. In the later works~\cite{Lee_2020_LearningQuadrupedal} and~\cite{Kumar_2021_RMA}, RL-trained policies made a quadrupedal robot traverse over various challenging terrains such as slippery ground, vegetation, and rocky terrain. They trained an encoder which compresses environmental information and enabled effective environment-aware locomotion. Moreover, Peng et al.~\cite{Peng_2020_LearningAgile} reproduced agile motions of animals by imitating recorded motion trajectory data. They made the Laikago robot walk and turn at moderate speed.

More complicated trained behaviors, such as dynamic recovery from a fall~\cite{Hwangbo_2019_LearningAgile, lee2019robust}, have been reported in the literature. These complex behaviors can be composed of a single framework using pre-trained expert networks and a gate neural network, and manifest agile and effective motions~\cite{Yang_2020_MultiExpert} on the real robot. Furthermore, RL can be utilized for bipedal robots to climb up stairs~\cite{Siekmann_2021_BlindBipedal} or to display diverse locomotion patterns such as standing, walking, and running~\cite{Siekmann_2021_SimtoReal}.

\begin{figure}[thpb]
\label{fig:Minicheetah}
\centering
\includegraphics[width=0.65\linewidth]{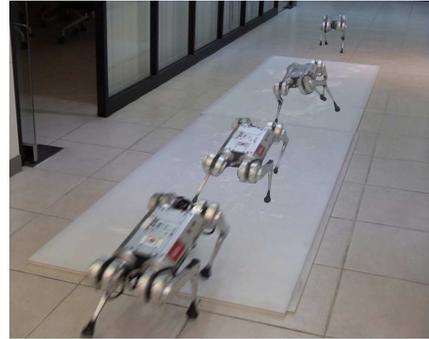}
    \caption{Dynamic and robust locomotion on a slippery plate of the friction coefficient of 0.22. The Mini Cheetah is traversing over it at the average speed of \unit[3.54]{m/s}.}
\vspace*{-0.7cm}
\end{figure}

The existing control approaches for quadrupedal locomotion rely on accurately estimated state input~\cite{Hwangbo_2019_LearningAgile, Lee_2020_LearningQuadrupedal, Kumar_2021_RMA, Yang_2020_MultiExpert, Gehring_2015_DynamicTrotting, Kim_2019_HighlyDynamic, Hong_2020_RealTime, Ding_2021_RepresentationFree}. However, we observed that existing state estimation algorithms become unreliable~\cite{Hartley_2019_IEKF, Kim_2021_LeggedRobot} on challenging terrains, such as ice and sand. Moreover, many state estimation algorithms, such as the one built in to the Mini Cheetah robot, require gait patterns a priori. Most neural network control policies often do not provide such information because the patterns are learned as well. Alternatively, contact states can be estimated either from dedicated contact sensors or from the model~\cite{Hwangbo_2016_ProbabilisticFoot}, but the former is computationally costly and the latter is prone to permanent damages during foot landing. Therefore, it is desirable to develop a control framework that does not rely on contact information. 

In addition, to walk and run on challenging terrains blind, information about the terrain must be estimated. An analytical method can be employed~\cite{Gehring_2015_DynamicTrotting} to estimate part of the information. Alternatively, it can be estimated implicitly using a trained neural network and proprioceptive state history~\cite{Lee_2020_LearningQuadrupedal, Kumar_2021_RMA}. The proprioceptive state history is useful for estimating both intrinsic robot states and extrinsic environment variables. However, this approach has two different major drawbacks. First, because the latent vectors are not interpretable, they cannot be used in conjunction with other modules that require state information. Second, the encoder training causes a significant computational overhead. An alternative to this approach is to directly estimate observable state variables such as the terrain angle. In our proposed approach, this information is indirectly estimated as a distance from the terrain to the foot.



To address the aforementioned shortcomings of the existing methods, we present a learning-based state estimation network, which is concurrently trained with the policy network. The efficacy of our method is demonstrated using the Mini Cheetah robot~\cite{Katz_2019_Minicheetah}, which is a lightweight and highly dynamic quadrupedal robot. Kim et al.~\cite{Kim_2019_HighlyDynamic} reports an MPC-based controller that could make Mini Cheetah run at up to \unit[3.7]{m/s} on a treadmill and ~\cite{Katz_2019_Minicheetah} achieves \unit[2.45]{m/s} in outdoor environments. We hereby report highly dynamic locomotion at \unit[3.74]{m/s} in various indoor and outdoor environments and robust and reliable locomotion behaviors on a slippery plate, bumpy asphalt road, and hills.

Our main contributions are as follows:
\begin{itemize}
    \item We propose a simple end-to-end locomotion learning framework that concurrently trains a control policy and a state estimator.
    \item Using the trained networks, we demonstrate dynamic locomotion on slippery terrains and slopes.
    \item We share the training details, such as the dynamic randomization and curriculum, so that our work can be reproduced by other researchers.
\end{itemize}


\section{METHOD}
Our goal is to develop an RL-based control framework that can follow the given velocity command, which consists of desired base linear velocities in the forward and lateral directions, and the desired yaw rate. We assume that the robot is equipped with an Inertial Measurement Unit (IMU) and joint encoders.

An overview of our control framework is illustrated in Fig.~\ref{fig:controlDiagram}. The framework consists of three different neural networks: the estimator, critic, and actor. The estimator network estimates multiple relevant state variables for control using the onboard sensors and feeds them to the actor network which outputs actuator commands. The critic network helps reduce variance in the policy gradient estimate from the RL algorithms. All neural networks are trained in simulation using RaiSim~\cite{Hwangbo_2018_PerContact}.

We use Proximal Policy Optimization (PPO)~\cite{Schulman_2017_PPO} for training the actor and critic, and supervised learning for training the estimator network. After a collection of a batch of trajectories in RaiSim, we update all three networks using their corresponding loss function. This process repeats like in the vanilla PPO until the performance metric converges. This concurrent learning of the two networks ensures that the policy network can adapt to the performance characteristics of the estimator network. For example, when the estimation is unreliable due to slippery foot contacts, the policy network will be trained with unreliable state estimates and manifest conservative behaviors.

The Mini Cheetah robot~\cite{Katz_2019_Minicheetah} is the robotic platform used in this work. Its compact size and powerful actuators enable us to tackle difficult tasks such as high-speed locomotion on slippery terrain. In addition, the source code for robot operation is available online\footnote{https://github.com/mit-biomimetics/Cheetah-Software}, which includes a high-performance locomotion controller~\cite{Kim_2019_HighlyDynamic}. Furthermore, its IMU sensor, 3DMGX5-AHRS made by Lord Corporation, provides not only the linear acceleration and angular velocity but also the estimated orientation based on the extended Kalman filter. Our version of the Mini Cheetah is \unit[1.8]{kg} heavier and \unit[1]{cm} longer compared to the original version presented in~\cite{Katz_2019_Minicheetah}. Our implementation code for the real robot can be found online\footnote{https://github.com/karlji1021/Cheetah-Software}.

\subsection{Training In Simulation}
The policy is trained on flat terrain in 800 different environments to efficiently collect samples. In each environment, the robot is initialized with highly random initial states as shown in the table~\ref{tab:intialState}. This helps the robot to recover from unexpected external disturbances such as interactions with humans or sudden changes in terrain parameters. With the probability of \unit[25]{\%}, the robot is initialized with the final state of the previous episode, whereby the robot can learn to overcome sudden changes in the velocity command in the real world. For further improvements, we also train the model in environments where an uneven flat terrain or slopes up to $\pm$\unit[10]{$^\circ$} are randomly generated. The uneven terrain was created using Perlin noise of the following parameters (fractal octaves = 5, fractal lacunarity = 3.0, fractal gain = 0.45, z-scale = $min$(0.21, 0.21$\cdot t^{-1}$) where $t$ is the number of iteration).

\begin{table}[thpb] 
\vspace*{-0.15 cm}
    \caption{Initial state noise}
    \label{tab:intialState}
    \begin{center}
        \begin{tabular}{|c|l|}
        \hline
        \multicolumn{1}{|c|}{\textbf{State}} & \multicolumn{1}{c|}{\textbf{Noise}}\\
        \hline
        Quaternion & $U^4(-0.2, 0.2)$, then normalized\\
        \hline
        Joint positions & \unit[$U^{12}(-0.2, 0.2)$]{rad}\\
        \hline
        Joint velocities & \unit[$U^{12}(-2.5, 2.5)$]{rad/s}\\
        \hline
        X linear velocity & \unit[$U^{1}(-1, 1)$]{m/s}\\
        \hline
        YZ linear velocities & \unit[$U^{2}(-0.5, 0.5)$]{m/s}\\
        \hline
        Angular velocities & \unit[$U^3(-0.7, 0.7)$]{rad/s}\\
        \hline
        \end{tabular}
    \end{center}
\vspace*{-0.35cm}
\end{table}

To train a policy more effectively, we set up a curriculum where the velocity command in the x-direction (i.e., forward/backward direction) gradually increases over each PPO iteration. At the early stage of training, the linear velocity command in the x-direction is uniformly sampled from \unit[U$^1(-0.5, 1.0)$]{m/s}. This range gradually enlarges up to \unit[U$^1(-1.75, 3.5)$]{m/s}, with the maximum forward command given according to 
\begin{align}
    V_{x,max} = 1 + {\frac{2.5}{1 + exp(-0.002\cdot(t - 1000))}},
\end{align}
where $t$ is the number of training iterations. Ten percent of the trajectories are then selected to have a zero velocity command for learning a standing still behavior.

We define reward functions and their coefficients as shown in the table~\ref{tab:reward}. The reward function is designed for two objectives: to follow the given command and to run in an efficient and natural way. The linear and angular velocity rewards are related to the former objective, and the other rewards are for the latter one. Most of the reward functions are shaped by referring to \cite{Hwangbo_2019_LearningAgile}. Among them, the foot clearance reward is important for a successful sim-to-real transfer because the relative foot positions to the ground and terrain geometry might be uncertain in some situations. The square-root function in the foot clearance reward is to increase its influence on the policy when the commanded velocity is too low. Airtime reward is designed for controlling swing-stance timing and generating standing still motions.

\begin{table}[thbp]
\vspace*{-0.15cm}
\caption{Reward Functions}
\vspace*{-0.25cm}
\label{tab:reward}
\begin{center}
\begin{tabular}{|p{0.97cm}|p{0.97cm}|p{0.97cm}|p{0.97cm}|p{0.97cm}|p{1.03cm}|}
\hline
\multicolumn{2}{|c|}{\textbf{Reward}}                                                                      & \multicolumn{4}{c|}{\textbf{Expression}}                                                                                                       \\ \hline
\multicolumn{2}{|c|}{Linear velocity}                                                            & 
    \multicolumn{4}{l|}{$r_v=k_{v}exp(-\vert\vert cmd_{v_{xy}}-V_{xy}\vert\vert ^2)$}        \\ \hline

\multicolumn{2}{|c|}{Angular velocity}                                                             &             \multicolumn{4}{l|}{$r_{\omega} = k_{\omega}exp(-1.5(cmd_{\omega_{z}}-\omega_{z}) ^2)$}     \\ \hline

\multicolumn{6}{|l|}{\ \ \ \ \ \ \ \ Airtime}                                                                                                                                                                 \\ \hline
\multicolumn{6}{|l|}{
    $r_{air,i} = 
    \begin{cases}
        \text{if stance cmd:}
        \qquad\qquad\ \; k_{a}clip(T_{s,i}-T_{a,i}, -0.3, 0.3) \\
        \text{else:}
        \begin{cases}
            \text{if} \  T_{max,i} < 0.25: \ 
            k_{a}min(max(T_{s,i}, T_{a,i}), 0.2)\\
            \text{else:}
            \qquad\qquad\;\;\;\;\; \ \ \ \ 0
        \end{cases}
    \end{cases}$
} \\ \hline

\multicolumn{2}{|c|}{Foot slip}                                                                    &             
    \multicolumn{4}{l|}{$r_{slip,i} = k_{slip}C_{f,i}\vert\vert V_{f,xy,i}\vert\vert ^2$}
                                                                                                   \\ \hline
\multicolumn{2}{|c|}{Foot clearance}                                                               &             
    \multicolumn{4}{l|}{$r_{cl,i} = k_{cl}( _{w}p_{f,z,i}-_{w}p^{des}_{f,z}) ^2\vert\vert V_{f,xy,i}\vert\vert ^{0.5}$}
                                                                                                   \\ \hline
\multicolumn{2}{|c|}{Orientation}                                                                 &             
    \multicolumn{4}{l|}{$r_{ori} = k_{ori} (angle(\phi_{body,z}, \phi_{world,z})) ^2$}
                                                                                                  \\ \hline
\multicolumn{2}{|c|}{Joint torque}                                                                 &             
    \multicolumn{4}{l|}{$r_{\tau} = k_{\tau}\vert\vert \tau\vert\vert ^2$}
                                                                                                 \\ \hline
\multicolumn{2}{|c|}{Joint position}                                                               &             
    \multicolumn{4}{l|}{$r_{q} = k_{q}\vert\vert q_t-q_{nominal}\vert\vert ^2$}
                                                                                               \\ \hline
\multicolumn{2}{|c|}{Joint speed}                                                                  &             
    \multicolumn{4}{l|}{$r_{\dot{q}} = k_{\dot{q}}\vert\vert \dot{q}_t\vert\vert ^2$}
                                                                                                   \\ \hline
\multicolumn{2}{|c|}{Joint acceleration}                                                           &             
    \multicolumn{4}{l|}{$r_{\ddot{q}} = k_{\ddot{q}}\vert\vert \dot{q}_t-\dot{q}_{t-1}\vert\vert ^2$}
                                                                                                  \\ \hline
\multicolumn{2}{|c|}{Action smoothness 1}                                                           &             
    \multicolumn{4}{l|}{$r_{s1} = k_{s1}\vert\vert q^{des}_t-q^{des}_{t-1}\vert\vert ^2$}
                                                                                                   \\ \hline
\multicolumn{2}{|c|}{Action smoothness 2}                                                           &             
    \multicolumn{4}{l|}{$r_{s2} = k_{s2}\vert\vert q^{des}_t-2q^{des}_{t-1}+q^{des}_{t-2}\vert\vert ^2$}
                                                                                                  \\ \hline
\multicolumn{2}{|c|}{Base motion}                                                                  &             
    \multicolumn{4}{l|}{$r_{base} = k_{base}(0.8V_z^2+0.2\vert \omega_x\vert  + 0.2\vert \omega_y\vert)$}
                                                                                                  \\ \hline
\multicolumn{6}{|c|}{\textbf{Reward Coefficients}}    \\ \hline                                

$k_v$      & 3.0     & $k_{cl}$            & -15.0            & $k_{\dot{q}}$      & -6e{-4} \\ \hline
$k_\omega$ & 3.0     & $k_{ori}$           & -3.0             & $k_{\ddot{q}}$    & -0.02          \\ \hline
$k_a$      & 0.3   & $k_\tau$            & -6e{-4} & $k_{s1}, k_{s2}$ & -2.5, -1.2     \\ \hline
$k_{slip}$ & -0.08 & $k_q$               & -0.75          & $k_{base}$       & -1.5           \\ \hline                                                                                                  
\end{tabular}
\end{center}
\vspace*{-0.25cm}
\end{table}

In the table~\ref{tab:reward}, $cmd$ is an abbreviation of command and $i$ is an index of the foot. $T_{a,i}$ and $T_{s,i}$ represent the time since last takeoff and touchdown, respectively, while being initialized to zero whenever a transition happens. $C_{f,i}$ in the foot slip reward is the contact state of each foot. In the foot clearance reward, $_{w}p^{des}_{f,z}$ is the desired foot height and it is set to \unit[0.09]{m}. We define a positive reward sum as $r_{pos}=r_v+r_\omega+\sum_{i=0}^3 r_{air,i}$ and a negative reward sum as $r_{neg}=\sum_{i=0}^3 (r_{slip,i}+r_{cl,i})+r_{ori}+r_{\tau}+r_{q}+r_{\dot{q}}+r_{\ddot{q}}+r_{s1}+r_{s2}+r_{base}$. The total reward is defined as
\begin{align}
    r_{tot}=r_{pos} \cdot exp(0.2r_{neg})
\end{align}
This form of a reward function ensures that the resulting reward is always positive and discourages the policy to choose an early termination.

Whenever the body of the robot except the knees and feet contacts the environment, the episode terminates and the robot is punished with a reward of -10. Therefore, the policy is trained toward reducing unnecessary collisions.

\subsection{Network Architecture}
\begin{figure*}[th]
\vspace*{0.25 cm}
\centering
\includegraphics[width=0.8\linewidth]{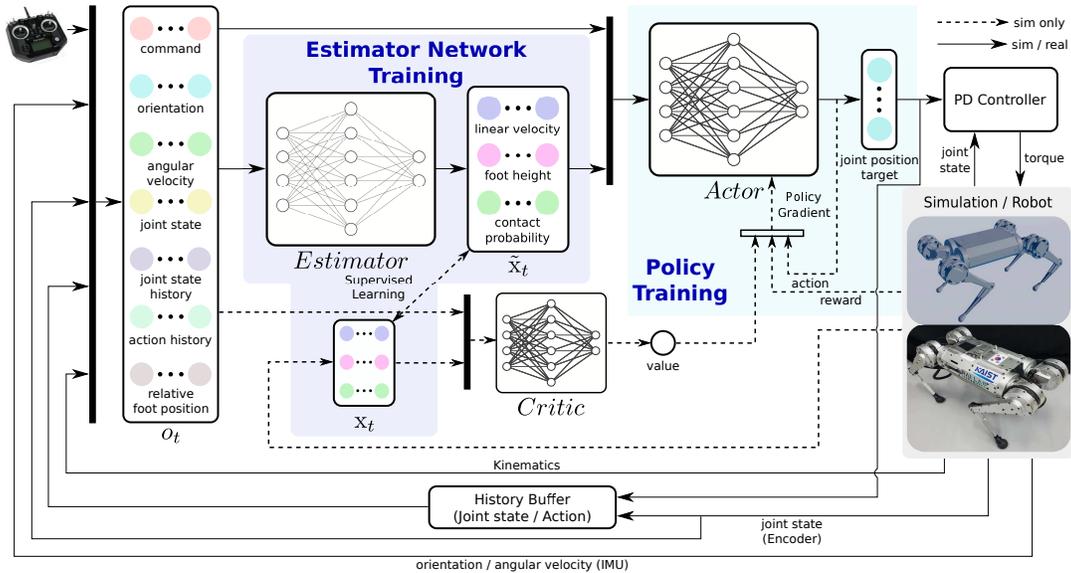}
    \caption{An overall control diagram and a proposed training framework for concurrent training of a control policy and a state estimator are shown. The estimator network takes sensor data $o_t$ as an input and outputs state variables, which are the base linear velocity, foot height, and contact probabilities. These estimated states are fed into the policy network together with the observation $o_t$. The estimator network is trained with supervised learning to reduce MSE between the estimated robot states and their corresponding true values obtained from simulation. An actor network is a policy which produces desired joint positions based on the state estimates. Both the critic and actor are trained using the PPO algorithm.}
    \label{fig:controlDiagram}
\vspace*{-0.45cm}
\end{figure*}

Our neural network structure consists of 3 components: an actor, a critic, and an estimator. All of them are designed as a Multi-Layer Perceptron~(MLP) network, with the actor and critic having a $[512\times256\times64]$ structure, and the estimator having a $[256\times128]$ structure. An MLP is the simplest neural network structure and computationally more efficient than other memory-based networks such as RNNs. We trained policies in a form of an LSTM but could not find meaningful differences in performance. The actor maps an observation to an action and the critic\cite{Sutton_2018_RL} estimates the value of the current state. The estimator network is to estimate states of the robot such as the base linear velocity. Those values are estimated by taking an observation $o_t$ as an input, and fed to the actor. The estimator network is trained using supervised learning with data from the simulation. Both the policy and the estimator are running synchronously at 100 Hz. The network structure is shown in Fig.~\ref{fig:controlDiagram}.

Our system takes sensor data as an input, and outputs desired joint positions for each actuator. Our framework still uses an analytical estimate of the gravity vector expressed in the body frame because it is computed by the IMU sensor. Furthermore, the estimation algorithms for the orientation are simple and reliable. Joint velocities are computed on the motor controllers by applying the finite difference method on joint positions.

The observation tuple is defined as
\begin{equation}\label{eq:observation}
    o_t=(\phi, \omega, q, \dot{q}, q^{des}_{t-1}, q^{des}_{t-2}, Q_{hist}, \dot{Q}_{hist}, _{b}p_{f}, cmd)
\end{equation}

\noindent where $\phi$ and $\omega$ are the base orientation and angular velocity, $q$ and $\dot{q}$ are the joint positions and velocities, $q^{des}_{t-1}$ and $q^{des}_{t-2}$ are the desired joint position targets for two previous time steps, $Q_{hist}$ and $\dot{Q}_{hist}$ are the joint position error history and joint velocity history, $_{b}p_{f}$ is the Cartesian positions of the feet relative to the center of mass expressed in the body frame, and $cmd$ is the given velocity command. The Cartesian foot positions are for observing where the feet are located, and it is known to be helpful for training a policy for complicated systems\cite{Reda_2020_LearningToLocomote}. For our study, joint state history is selected at $t-\unit[0.02]{s}$, $t-\unit[0.04]{s}$, and $t-\unit[0.06]{s}$. For stable learning and control, all observation variables are normalized to have a mean of 0.0 and a standard deviation of 1.0. For the same reason, policy outputs are multiplied by a nominal value and then added to nominal joint positions to obtain the desired joint target distributions. This relationship is expressed as
\centerline{$q^{des}_t=q_{nominal}+\sigma_{a}a_t$,}
\noindent where $q_{nominal}$ is the nominal joint positions, which is the same as the standing up configuration, $\sigma_a=0.1$ is a predefined action scaling factor, and $a_t$ is an output of the policy network. The desired joint positions are computed at \unit[100]{Hz} and converted to joint torques by a PD controller module with $K_p$=\unit[17]{N$\cdot$m$\cdot$rad$^{-1}$} and $K_d$=\unit[0.4]{N$\cdot$m$\cdot$s$\cdot$rad$^{-1}$}, at \unit[40]{kHz} on the real robot.

The estimator network is designed to predict the state of the robot without utilizing a dedicated estimation algorithm. In this paper, the linear velocity, foot height, and contact probability are estimated. The linear velocity estimate is essential in the following velocity command. By removing the necessity of sophisticated state estimation algorithms, the implementation on the robot becomes much simpler. It also has an advantage that the controllers become robust against inevitable errors of the state estimator. As illustrated in \cite{Kim_2021_LeggedRobot}, estimation of the linear velocity under highly erroneous environments is vulnerable to foot slips. Our end-to-end neural network structure avoids such a challenge in two ways. First, the estimator network is trained in environments where the feet slip often. Therefore, it can still produce a reasonable estimate of the linear velocity using other sensor information and previous observations. Second, the policy network is trained with imperfect information such that it is aware of possible slippages.

The idea behind learning foot height and contact probability is to achieve the sufficient foot clearance. Due to the wide range of velocity command and the clearance reward that penalizes high speed, the foot clearances become smaller at low speeds. Foot clearance plays an important role in a sim-to-real transfer because insufficient foot clearance might lead to an early foot landing or tripping while running. We discovered that the reward term alone is not sufficient to learn to maintain sufficient foot clearances. Our solution to this problem is to estimate the foot clearances and feed them directly to the policy network. We note that a learning-based estimator is capable of approximately estimating the foot clearance from the observation. First, foot contact states are obtainable from joint position errors. Second, a terrain slope becomes observable from the foot contact states, orientation, and joint positions. Therefore, as the slope is observable, the estimator network can compute the foot height under the assumption that the terrain is even.

\subsection{Dynamics Randomization}
Dynamics randomization is important for a successful sim-to-real transfer of policies trained in simulation~\cite{Peng_2018_SimtoReal}. In our case, the robot controller learned without dynamics randomization exhibits shaky motions when deployed on the real robot. It comes from the fact that kinematic and dynamic parameters such as leg length, actuator positions, and center of mass, are not exactly the same as those used in the simulation. Consequently, this reality gap often makes the robot unable to reach sufficient performance. To eliminate this gap, we randomize several components as follows:

\begin{itemize}
    \item observation noise
    \item motor frictions
    \item PD controller gains
    \item foot positions and collision geometry
    \item ground friction
\end{itemize}
These parameters are randomized at the start of each episode or iteration.

The observation noise is added during the training phase in the simulation. On the real robot, joint velocity measurement comes from numerical differentiation of the joint positions, which causes errors in the joint velocity observation. Moreover, fluctuation in the logging frequency might lead to a failure in updating the velocity values for a single time step. Such an event corresponds to a delay of \unit[2]{milliseconds}. Therefore, the joint position and velocity measurements in simulation are randomized to reflect the true distribution of the measurements; they are sampled from \unit[U$^{12}$(-0.05, 0.05)]{rad} and \unit[U$^{12}$(-0.5, 0.5)]{rad/s}, respectively. For the same reason, a uniformly distributed noise U$^4$(-0.03, 0.03), \unit[U$^4$(-0.03, 0.03)]{m}, and \unit[U$^3$(-0.1, 0.1)]{rad/s} are added to the base orientation, foot position, and base angular velocity, respectively.

Motor friction is randomized to reflect the differences between actuator units. We chose a conservative range of \unit[U$^1$(0, 0.3)]{N$\cdot$m} for the hip abduction/adduction~(HAA) and hip flexion/extension~(HFE), and \unit[U$^1$(0.1, 0.7)]{N$\cdot$m} for the knee flexion/extension~(KFE). Their measured friction values on the real robot are 0.2, 0.2, and \unit[0.5]{N$\cdot$m} for HAA, HFE, and KFE, respectively. The KFE joints have higher friction because they have an extra belt transmission. The PD controller gains are randomized to mitigate the effects of motor friction and damping. We added a uniform noise of \unit[U$^1$(-2, 2)]{N$\cdot$m$\cdot$rad$^{-1}$}, and \unit[U$^1$(-0.1, 0.1)]{N$\cdot$m$\cdot$s$\cdot$rad$^{-1}$} for the position and velocity gains, respectively.

The foot position and collision geometry are randomized to reduce both effects of measurement errors and the deformation of the rubber feet. The foot position observations are disturbed with a uniform noise of \unit[U$^1$(-10, 10)]{mm} in the longitudinal direction, \unit[U$^1$(-5, 5)]{mm} in the lateral direction, and \unit[U$^1$(-20, 20)]{mm} in the leg length direction. These noises are added to the measured foot positions. The foot sphere radii are randomized to \unit[U$^1$(6, 10)]{mm}.

Finally, the ground friction was randomized to U$^1$(0.4, 1.0). Owing to this randomization, the robot can run not only on very slippery ground but also on the ground with very high friction like asphalt.

\section{RESULTS}
Part of the results in this section can also be found in the accompanying video.
\subsection{Controller Descriptions}
To evaluate the performance of the proposed network structures and analyze how each component affects different performance metrics, we test the following settings:

\begin{itemize}
    \item \textit{Implicit}: As a baseline, the explicit state estimator in our proposed framework is substituted with an implicit estimator as in ~\cite{Lee_2020_LearningQuadrupedal, Kumar_2021_RMA}.
    \item \textit{Sequential}: In phase 1, a policy is trained with the ground truth robot states. In phase 2, a state estimator was trained with the final policy of phase 1. For implementation, the estimator replaces the ground truth input.
    \item \textit{Built-in MPC}: The MPC controller described in \cite{Katz_2019_Minicheetah}
    \item \textit{RL-LKF}: An RL policy in a single MLP form trained with the linear velocity data from the simulator and uses an LKF state estimator on the real robot.
    \item \textit{Concurrent}: Our proposed control framework trained on relatively flat ground
    \item \textit{Concurrent+Slope}: Our proposed control framework trained on randomly generated slopes.
\end{itemize}

\noindent In addition to the above models, we also created four other network models by excluding one of the three estimated states or all of them (i.e., \textit{w/o Linvel Estimator}, \textit{w/o FootHeight Estimator}, \textit{w/o Contact Estimator}, \textit{w/o Estimator}) for ablation studies in simulation.

For comparison of the explicit and implicit estimators, we trained the \textit{Implicit} model. The implicit estimator follows the framework in~\cite{Kumar_2021_RMA}. During the phase 1, the encoder takes the observation $o_t \texttt{\textbackslash} (Q_{hist}, \dot{Q}_{hist})$, linear velocity, foot height, and foot contact as an input. For the adaptation module, the history length of 20 is used and the network consists of 3 layers of 1D CNN. The output dimension of the encoder and adaptation module is 11. All the other settings are the same as our framework. The total training time is 7 hours for phases 1 and 2 altogether.

\subsection{Evaluation of the Performance in Simulation}
\subsubsection{Learning Performance}
\begin{figure}[thpb]
  \vspace*{0.25cm}
\centering
\includegraphics[width=\linewidth]{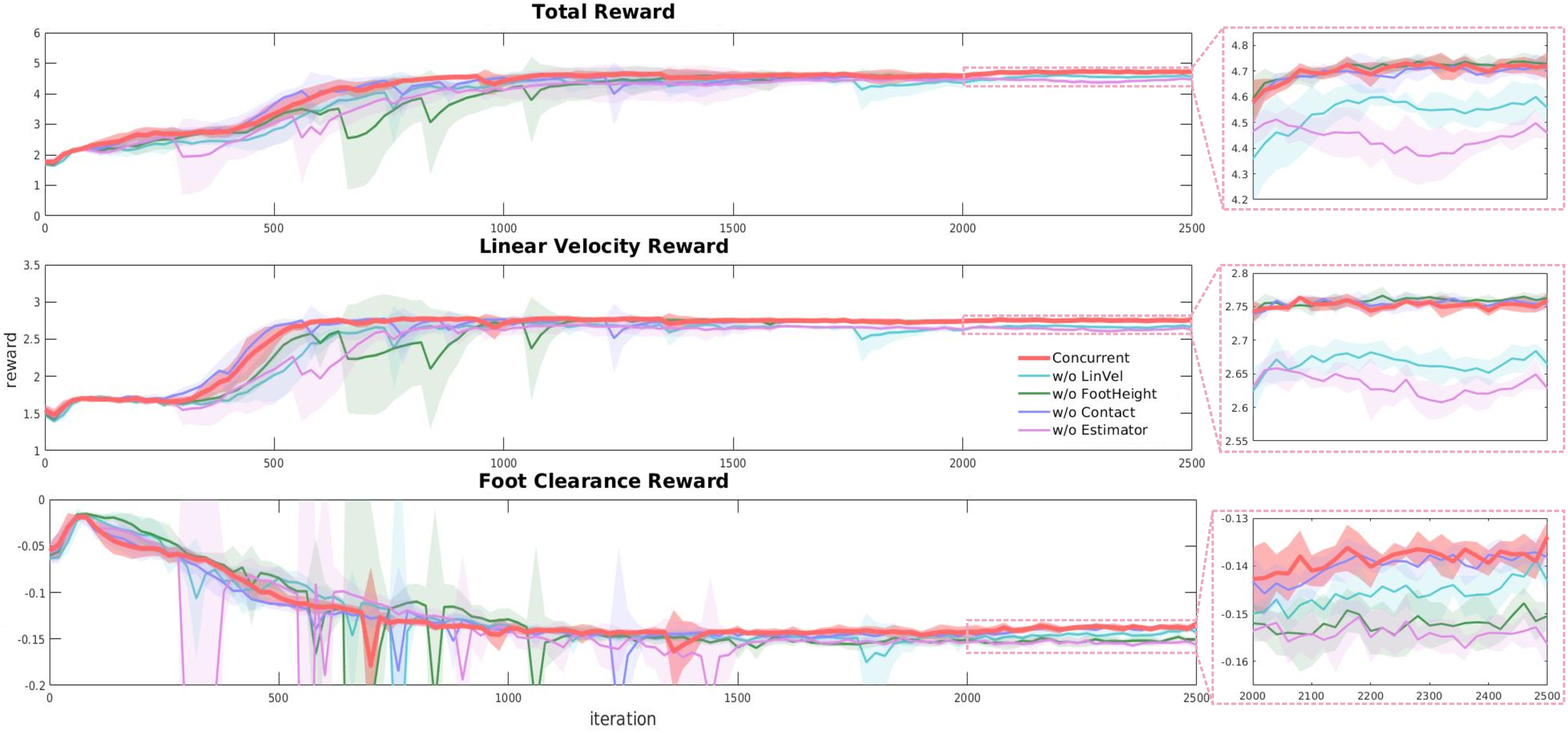}
    \caption{Learning curves of the total reward, linear velocity reward, and foot clearance reward are shown. The linear velocity, foot height, and contact probability are estimated. The \textit{Concurrent} model has the highest performance and learning stability, while the model without an estimator~(\textit{w/o Estimator}) has the lowest performance.}
    \label{fig:learningCurve}
\vspace*{-0.15 cm}
\end{figure}

First, we compared the performance of the learned controllers (i.e., \textit{Concurrent}, \textit{w/o Linvel Estimator}, \textit{w/o FootHeight Estimator}, \textit{w/o Contact Estimator}, and \textit{w/o Estimator}) in simulation. Each network structure is trained four times and their average learning curves are shown in Fig.~\ref{fig:learningCurve}. All the models were trained until they converged to a stable expected return. After 2500 iterations, which consumed about 800 million samples and 4 hours of training in real-time, they all converged to a stable value. The rewards of the trained models are summarized in the table~\ref{tab:learning}.

As shown in Fig.~\ref{fig:learningCurve} and the table~\ref{tab:learning}, the \textit{Concurrent} model converges to the highest rewards while the \textit{w/o Estimator} and \textit{Implicit} models converge to the lowest. Out of the three estimated states, linear velocity estimation plays the most important role in improving the policy. Omitting the linear velocity estimation leads to a significant drop in metrics: the total, linear velocity, and foot clearance rewards. This result proves that the linear velocity is crucial for learning the locomotion of legged robots.

Another noticeable improvement comes from the foot height estimation. Explicitly estimating the foot height improves the foot clearance of the robot, resulting in a higher foot clearance reward. The effectiveness of the higher foot clearance will be discussed in the \textit{Locomotion on rough terrains} section.

Foot contact probability estimation makes the least impact on the final performance, but it stabilizes and accelerates learning processes as shown in the total reward graph in Fig.~\ref{fig:learningCurve}.

\subsubsection{Tracking and Estimation Error}
For further investigation on the impact of the estimator network, we tested the \textit{Concurrent}, \textit{w/o Estimator}, and \textit{Implicit} models in simulation. All models were given the same random commands every 20 seconds and for 10 minutes in total while the friction coefficient of the flat ground is kept at 0.6. The performance was only measured for the steady-state errors. The velocity commands are sampled from \unit[U$^1$(-1.75, 3.5)]{m/s}, \unit[U$^1$(-1, 1)]{m/s}, and \unit[U$^1$(-2, 2)]{rad/s}, for $V_x, V_y,$ and $\omega$, respectively.

The result is shown in the table~\ref{tab:ErrorInSim}. The \textit{Concurrent} model has the smallest RMS errors for following the given linear velocity. It means that the estimated states help the robot to stabilize itself. Furthermore, the tracking errors of the \textit{Implicit} model are on a similar level to that of \textit{w/o Estimator}. From this fact, we suggest that tracking performance does not benefit a lot from utilizing the implicit estimator. Models with the implicit estimator having a latent vector of sizes 8 and 20 showed worse performance than the presented one.

Interestingly, our concurrently trained model performs better than a sequentially trained model. From the table~\ref{tab:ErrorInSim}, the \textit{Sequential} model has slight performance degradation. We hypothesize that this is because the policy trained with a state estimator tends to avoid states where the state estimator becomes unreliable. This problem can be easily solved by training them concurrently as we proposed. Note also that the concurrent training requires only one training dataset, which is more efficient than the sequential training.

\subsubsection{Locomotion on rough terrains}
We investigated the effectiveness of the foot clearance of the learned models. The foot clearance should be sufficiently high for traversing over the rough terrains. Therefore, in this experiment, we compare the average time to fall on the rough terrains with z-scale of 0.525. Commands are sampled from \unit[U$^1$(-1, 1)]{m/s}, and \unit[U$^1$(-1, 1)]{rad/s} where the foot clearance is relatively small.

From the table~\ref{tab:ErrorInSim}, the \textit{Concurrent} model shows an overwhelming performance compared to the other models. It is robust against a fall due to increased foot clearance. On the other hand, an implicit estimator and a single policy do not have sufficient foot clearance. Eventually, they are easy to fall and have worse locomotion performance. In conclusion, we suggest that the explicit estimation of the foot height significantly contributes to locomotion performance.

\subsubsection{Computational Cost}
The trained estimator has computational benefits over analytical state estimators. Using a single core of Ryzen9 5950x, the estimator network takes \unit[7]{$\mu$s} for a forward pass, while the linear Kalman filter on the Mini Cheetah consumes \unit[34]{$\mu$s}. Also, the implicit estimator with 20 history inputs takes \unit[20]{$\mu$s} which is about three times longer than the simple explicit estimator.

\begin{table}[t]
\vspace*{0.25 cm}
    \caption{Ablation Study for the Estimator:  \hspace{\textwidth} Reward of the learned models}
    \label{tab:learning}
    \begin{center}
        \begin{tabular}{|c|c|c|c|}
        \hline
        \multirow{3}{*}{\textbf{Model}} & \multicolumn{3}{c|}{\textbf{Reward}}\\
        \cline{2-4}
         & Total & \makecell{Linear \\ Velocity} & \makecell{Foot \\ Clearance}\\
        \hline
        \textit{Concurrent} & 4.7212 & \textbf{2.7595} & \textbf{-0.1338} \\
        \hline
        \textit{w/o LinVel Estimator} & 4.5555 & 2.6643 & -0.1432\\
        \hline
        \textit{w/o FootHeight Estimator} & \textbf{4.7293} & \textbf{2.7625} & -0.1504\\
        \hline
        \textit{w/o Contact Estimator} & 4.7125 & 2.7543 & -0.1382\\
        \hline
        \textit{w/o Estimator} & 4.4577 & 2.6284 & -0.1565\\
        \hhline{|=|=|=|=|}
        \textit{Implicit} & 4.439 & 2.611 & -0.1447\\
        \hline
        \end{tabular}
    \end{center}
\vspace*{-0.6 cm}
\end{table}

\begin{table}[thpb]
\caption{Tracking and Estimation Error \hspace{\textwidth} \& Locomotion on Rough Terrains}
\label{tab:ErrorInSim}
\begin{center}
\begin{tabular}{|c|c|c|c|c|}
\hline
\textbf{Task}                                & \multicolumn{1}{c|}{\textbf{Model}} & $V_x$ [m/s] & $V_y$ [m/s] & $\omega_z$ [rad/s] \\ \hline
\multirow{4}{*}{\makecell{\textbf{Command} \\ \textbf{Following}}} & \textit{Concurrent} & \textbf{0.1112} & \textbf{0.0708} & 0.1222     \\ \cline{2-5} 
& \textit{w/o Estimator}  & 0.1725 & 0.0959 & \textbf{0.1137}     \\ \cline{2-5} 
& \textit{Implicit}  & 0.1679 & 0.1233 & 0.1379    \\ \cline{2-5} 
& \textit{Sequential}  & 0.1358 & 0.0996 & 0.1201    \\ \hline 
\multirow{2}{*}{\makecell{\textbf{Estimation} \\ \textbf{Error}}} & \textit{Concurrent} & \textbf{0.0243} & \textbf{0.0199} & - \\ \cline{2-5} 
& \textit{Sequential} & 0.06 & 0.036 & - \\ \hline
 & & \multicolumn{3}{c|}{Average time to fall [sec]} \\ \hline
\multirow{4}{*}{\textbf{\makecell{\textbf{Rough} \\ \textbf{Terrain}}}} & \textit{Concurrent} &  \multicolumn{3}{c|}{\textbf{85.7}}     \\ \cline{2-5} 
& \textit{w/o Estimator} & \multicolumn{3}{c|}{25.0}     \\ \cline{2-5} 
& \textit{Implicit} & \multicolumn{3}{c|}{30.0}    \\ \cline{2-5} 
& \textit{Sequential} & \multicolumn{3}{c|}{75.0}    \\ \hline
\end{tabular}
\end{center}
\vspace*{-0.5 cm}
\end{table}

\subsection{Evaluation of the Performance on the Real Robot}
We evaluated the performance of controllers on the real robot in terms of command following, state estimation, the maximum running speed, the maximum traversable slope angle, and foot clearance. The \textit{Concurrent+Slope} model requires 5000 iterations for convergence due to the challenging terrains. The summary of the performance is shown in the table~\ref{tab:Performance}.

\begin{table*}[thpb] 
\vspace*{0.25 cm}
    \caption{Performance of Controllers on the Real Robot}
    \label{tab:Performance}
    \begin{center}
        \begin{tabular}{|c|c|c|c|c|c|c||c|}
        \hline
         \multirow{2}{*}{\textbf{Task}} & \multirow{2}{*}{\makecell{\textbf{Terrain} \\ (friction)}} & \multirow{2}{*}{\textbf{Command}} & \multicolumn{5}{c|}{\textbf{Model}} \\
        \cline{4-8}
        & & & {\textit{Built-in MPC}} & {\textit{RL-LKF}} & {\textit{w/o Estimator}} & {\textit{Concurrent}} & {\textit{Concurrent+Slope}}\\
        \hline
        \multirow{6}{*}{\makecell{\textbf{Command Following} \\ \textbf{: Tracking Error}}} & \multirow{3}{*}{\makecell{Normal \\ ($\mu=0.6\text{-}0.88$)}} & $V_x$ [m/s] & 0.3041 & 0.2744 & 0.2635 & \textbf{0.2387} & 0.2488 \\
        \cline{3-8}
        & & $V_y$ [m/s] & - & 0.2031 & 0.2334 & \textbf{0.1722} & 0.1817 \\
        \cline{3-8}
        & & $\omega_z$ [rad/s] & 0.462 & 0.2857 & 0.2415 & 0.2427 & \textbf{0.2395} \\
        \cline{2-8}
        & \multirow{3}{*}{\makecell{Slippery \\ ($\mu=0.22$)}} & $V_x$ [m/s] & - & 0.3255 & 0.2888 & 0.2874 & \textbf{0.2532} \\
        \cline{3-8}
        & & $V_y$ [m/s] & - & 0.285 & 0.2617 & \textbf{0.2183} & 0.2446 \\
        \cline{3-8}
        & & $\omega_z$ [rad/s] & 0.525 & 0.3229 & 0.2709 & \textbf{0.2504} & 0.2654 \\
        \hline
        \multirow{4}{*}{\makecell{\textbf{State Estimation} \\ \textbf{: Estimation Error}}} & \multirow{2}{*}{\makecell{Normal \\ ($\mu=0.6\text{-}0.88$)}} & $V_x$ [m/s] & 0.1869 & 0.0808 & \textbf{0.0653} & 0.0783~(0.1069) & 0.0852~(0.0946) \\
        \cline{3-8}
        & & $V_y$ [m/s] & - & 0.1115 & 0.0666 & 0.1227~(0.0793) & \textbf{0.0557}~(0.078) \\
        \cline{2-8}
        & \multirow{2}{*}{\makecell{Slippery \\ ($\mu=0.22$)}} & $V_x$ [m/s] & - & 0.1575 & 0.0741 & \textbf{0.0533}~(0.1209) & 0.1168~(0.1029) \\
        \cline{3-8}
        & & $V_y$ [m/s] & - & 0.0623 & 0.0612 & \textbf{0.0555}~(0.0848) & 0.0833~(0.1061) \\
        \hline
        \multirow{2}{*}{\makecell{\textbf{Maximum} \\ \textbf{Average Speed}}} & \multicolumn{2}{c|}{Normal ($\mu=0.6\text{-}0.88$) [m/s]} & 1.72 & 2.18 & 3.20 & 3.33 & \textbf{3.75} \\
        \cline{2-8}
        & \multicolumn{2}{c|}{Slippery ($\mu=0.22$) [m/s]} & 1.34 & 2.19 & 3.14 & 3.25 & \textbf{3.54} \\
        \hline
        \textbf{Maximum Slope} & \multicolumn{2}{c|}{\makecell{Normal ($\mu=0.6\text{-}0.88$) / \\ Slippery ($\mu=0.22$) [$^{\circ}$]}} & 12.4 & 12.4 & 9.6 & 12.4 & \textbf{19.1} / 9.0 \\
        \hline
        \makecell{\textbf{Maximum} \\ \textbf{Foot Height}} & \multicolumn{2}{c|}{Normal [cm]} & \textbf{5} & 2 & 2 & 3 & 3 \\
        \hline
        \end{tabular}
    \end{center}
\vspace*{-0.6 cm}
\end{table*}

\subsubsection{Network Implementation on the Real Robot}
For the experiments in the real environments, we compared the five aforementioned control models: \textit{Built-in MPC}, \textit{RL-LKF}, \textit{Concurrent}, and \textit{Concurrent+Slope}. As described in the \textit{Controller Description} section, the \textit{RL-LKF} model requires an LKF state estimator. However, we could not use the built-in contact estimator as it assumes a periodic gait schedule, while our learned policy inherently changes gait patterns over speeds. Therefore, when estimating the contact state of the \textit{RL-LKF} model for the LKF, we used the proprioceptive touchdown detection method described in~\cite{Hyun_ImplementationOfTrot}. When the difference between the KFE joint position and its desired position is over the threshold of \unit[-0.4]{rad}, we assume that the leg is in contact.

\subsubsection{Command Following and State Estimation}
In the real environments, each controller was tested on normal and slippery ground with a step velocity command of \unit[1.0]{m/s}, \unit[0.8]{m/s}, and \unit[2.0]{rad/s} for linear velocities in x and y directions, and a yaw rate, respectively. The commands continued for 1 second and the robot started from zero commands. The slippery plate covered with boric acid powder has a friction coefficient of 0.22, which is much lower than that of the training environments. The results are shown in the table~\ref{tab:Performance}. In some cases, the \textit{Built-in MPC} controller fell and those instances are marked with '-' in the table. On the other hand, RL policies performed robustly against all sudden step inputs in this experiment. The \textit{Concurrent} model has the best tracking performance, while the \textit{Concurrent+Slope} model has a slightly higher error possibly due to the fact that the \textit{Concurrent} model is overfitted to simpler terrains.

We also recorded the estimation data from the LKF while the \textit{Concurrent} and \textit{Concurrent+Slope} models are running. The errors from the estimator networks are written in the parentheses. We note that estimation error does not necessarily lead to a deterioration in tracking performance. We suppose that other observation inputs, such as joint state history, mitigate the effects of the estimation errors so that the concurrent learning framework becomes robust against these errors.

\subsubsection{High-Speed Locomotion}
The \textit{Concurrent+Slope} model has the highest maximum speed. We tested each controller repeatedly until the robot fell. The maximum outdoor speed of our model, \unit[3.75]{m/s}, is comparable to the one reported by Kim et al.~\cite{Kim_2019_HighlyDynamic}, who report outdoor speed of over \unit[1.7]{m/s} and treadmill speed of \unit[3.7]{m/s}.

Our \textit{Concurrent+Slope} controller is capable of running at \unit[3.54]{m/s} on a slippery plate with $\mu=0.22$. When large foot slippages occurred, the robot recovered quickly, even when the robot is running near the maximum speed. If the command is suddenly set to zero while running, the robot makes a stable pose to stop quickly. We assume that this high performance is achieved owing to two factors: the policy trained on low friction terrains and its independence on a state estimation algorithm. Because our proposed framework is trained to be aware of possible foot slippages, it can be robust against estimation errors.

On both normal and slippery ground, the \textit{Concurrent} model exhibits better performance than the \textit{Built-in MPC} and \textit{RL-LKF} models. The \textit{Built-in MPC} model could not reach speeds over \unit[1.7]{m/s} on the normal ground and was easy to fall on the slippery ground at speed under \unit[1.3]{m/s}. \textit{RL-LKF} model also runs at rather conservative speeds lower than \unit[2.2]{m/s}. On the other hand, all the RL controllers are able to run consistently on all tested terrains as they are trained on terrains with the various friction coefficients.

\subsubsection{Locomotion on Hills}
Training a policy on slopes with random angles and friction coefficients significantly improves climbing performance. The \textit{Concurrent+Slope} model is capable of climbing a normal hill up to \unit[19.1]{$^\circ$}, which is steeper than slope angles of the training environments, $\pm$\unit[10]{$^\circ$}. Also, we note that \textit{Concurrent+Slope} model can walk up a slippery hill up to \unit[9.0]{$^\circ$}. Although the feet of the robot are slipping on it, the robot managed to climb up the slope by pushing the ground with adequate force. This behavior is partially learned in simulation, but the robot adapts its motion for the much more slippery slope of the friction coefficient of 0.22. For outdoor environments, we demonstrate the performance of our policy on a bumpy and hilly asphalt road. Please refer to the supplementary video for the demonstration.

The other controllers could not climb up a normal hill with angles steeper than \unit[12.4]{$^\circ$}. Because the RL controllers except for the \textit{Concurrent+Slope} model are trained only on nearly flat terrains, they have unsatisfactory climbing performance. Although the other controllers' maximum traversable slope angles are on a similar level, they display different locomotion characteristics. The \textit{Built-in MPC} model has higher foot clearances, but its non-stopping gait makes the robot unstable. The other RL models show relatively lower foot clearances, while their standing-still behavior significantly improves the stability of the robot.

\subsubsection{Foot Clearance}
We evaluated the foot clearance of each controller while the robot is running at \unit[1.0]{m/s}. The maximum foot clearance is shown in the table~\ref{tab:Performance}. We could recognize that foot clearance of the \textit{Concurrent} and \textit{Concurrent+Slope} models on the real robot is higher than the models without the estimator network, \textit{RL-LKF} and \textit{w/o Estimator}. As shown in the simulation test, lower foot clearance hinders stable locomotion on highly uneven terrains. This issue remains equally problematic on the real robot. We discovered that the \textit{RL-LKF} and \textit{w/o Estimator} models exhibit lower foot clearance at low speeds. Therefore, we suggest that explicit estimation of the foot clearance is effective for improving the performance.

\subsubsection{Contact Estimation}
\begin{figure}[thpb]
\vspace*{-0.15cm}
\centering
\includegraphics[width=\linewidth]{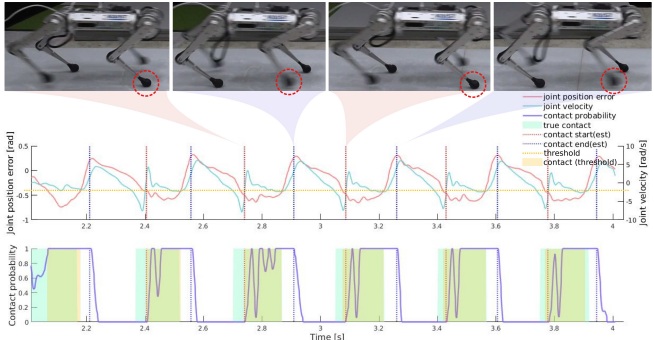}
    \caption{A contact estimation diagram for the front right foot is shown. KFE joint position error, joint velocity, and contact state are drawn. Contact starts with an abrupt change of the joint velocity by impact with the ground and ends with upward joint motion.}
\vspace*{-0.5cm}
\label{fig:ContactEstimation}
\end{figure}

Our proposed estimator network outputs contact probability for each foot. In Fig.~\ref{fig:ContactEstimation}, the contact state probability of the front right foot is drawn with the real contact state, KFE joint position error and joint velocity. The estimated contacts are shifted by 0.04 seconds from the actual contacts, which corresponds to 3 control steps excluding 0.01 seconds of communication delay. This is because there is a reality gap due to the compliance of the chain and the rubber feet of the real robot. In addition, the joint history inputs are sparsely sampled with 0.02-second intervals and it introduces further delay in detection. The estimator network detects a contact when the joint abruptly stops by impacts with the ground. The diagram also justifies the threshold of -0.4 rad of joint position error for the contact estimation used for LKF.

\section{CONCLUSION}
We presented a framework for concurrent training of a control policy and a state estimator. This framework requires neither an advanced control algorithm nor an accurate state estimation algorithm. Therefore, it requires significantly less effort for implementation on the real-legged system. Furthermore, our concurrent training method outperforms implicit estimation methods and a sequential training method in many aspects such as command tracking, robustness on rough terrain, and training time. To the best of our knowledge, our record is the fastest reported legged locomotion using reinforcement learning. The robot is also able to stably run on a slippery plate even under foot slippages. Although foot slippages often compromise the quality of the state estimation, the concurrently trained policy is robust against such issues.

The proposed learning-based state estimation can be useful without the control policy in many applications. It can provide reliable state estimates for motion analysis. Furthermore, we expect that the interpretable state outputs from our proposed network can be useful in conjunction with other controllers, such as MPC-based ones. 




\bibliographystyle{IEEEtran}
\bibliography{IEEEabrv, references}

\begin{thebibliography}{10}
\providecommand{\url}[1]{#1}
\csname url@samestyle\endcsname
\providecommand{\newblock}{\relax}
\providecommand{\bibinfo}[2]{#2}
\providecommand{\BIBentrySTDinterwordspacing}{\spaceskip=0pt\relax}
\providecommand{\BIBentryALTinterwordstretchfactor}{4}
\providecommand{\BIBentryALTinterwordspacing}{\spaceskip=\fontdimen2\font plus
\BIBentryALTinterwordstretchfactor\fontdimen3\font minus
  \fontdimen4\font\relax}
\providecommand{\BIBforeignlanguage}[2]{{%
\expandafter\ifx\csname l@#1\endcsname\relax
\typeout{** WARNING: IEEEtran.bst: No hyphenation pattern has been}%
\typeout{** loaded for the language `#1'. Using the pattern for}%
\typeout{** the default language instead.}%
\else
\language=\csname l@#1\endcsname
\fi
#2}}
\providecommand{\BIBdecl}{\relax}
\BIBdecl

\bibitem{Hwangbo_2019_LearningAgile}
J.~Hwangbo, J.~Lee, A.~Dosovitskiy, D.~Bellicoso, V.~Tsounis, V.~Koltun, and
  M.~Hutter, ``Learning agile and dynamic motor skills for legged robots,''
  \emph{Science Robotics}, vol.~4, no.~26, 2019.

\bibitem{Hutter_2016_ANYmal}
M.~Hutter, C.~Gehring, D.~Jud, A.~Lauber, C.~D. Bellicoso, V.~Tsounis,
  J.~Hwangbo, K.~Bodie, P.~Fankhauser, M.~Bloesch, R.~Diethelm, S.~Bachmann,
  A.~Melzer, and M.~Hoepflinger, ``Anymal - a highly mobile and dynamic
  quadrupedal robot,'' in \emph{2016 IEEE/RSJ International Conference on
  Intelligent Robots and Systems (IROS)}, 2016, pp. 38--44.

\bibitem{Lee_2020_LearningQuadrupedal}
J.~Lee, J.~Hwangbo, L.~Wellhausen, V.~Koltun, and M.~Hutter, ``Learning
  quadrupedal locomotion over challenging terrain,'' \emph{Science Robotics},
  vol.~5, no.~47, 2020.

\bibitem{Kumar_2021_RMA}
A.~Kumar, Z.~Fu, D.~Pathak, and J.~Malik, ``Rma: Rapid motor adaptation for
  legged robots,'' in \emph{Robotics: Science and Systems Conference (RSS
  2021)}, 2021.

\bibitem{Peng_2020_LearningAgile}
X.~B. Peng, E.~Coumans, T.~Zhang, T.-W.~E. Lee, J.~Tan, and S.~Levine,
  ``Learning agile robotic locomotion skills by imitating animals,'' in
  \emph{Robotics: Science and Systems (RSS 2020)}, 07 2020.

\bibitem{lee2019robust}
J.~Lee, J.~Hwangbo, and M.~Hutter, ``Robust recovery controller for a
  quadrupedal robot using deep reinforcement learning,'' \emph{arXiv preprint
  arXiv:1901.07517}, 2019.

\bibitem{Yang_2020_MultiExpert}
C.~Yang, K.~Yuan, Q.~Zhu, W.~Yu, and Z.~Li, ``Multi-expert learning of adaptive
  legged locomotion,'' \emph{Science Robotics}, vol.~5, no.~49, 2020.

\bibitem{Siekmann_2021_BlindBipedal}
J.~Siekmann, K.~Green, J.~Warila, A.~Fern, and J.~Hurst, ``Blind bipedal stair
  traversal via sim-to-real reinforcement learning,'' in \emph{Robotics:
  Science and Systems Conference (RSS 2021)}, 2021.

\bibitem{Siekmann_2021_SimtoReal}
J.~Siekmann, Y.~Godse, A.~Fern, and J.~Hurst, ``Sim-to-real learning of all
  common bipedal gaits via periodic reward composition,'' in \emph{2021 IEEE
  International Conference on Robotics and Automation (ICRA)}, 2021.

\bibitem{Gehring_2015_DynamicTrotting}
C.~Gehring, C.~D. Bellicoso, S.~Coros, M.~Bloesch, P.~Fankhauser, M.~Hutter,
  and R.~Siegwart, ``Dynamic trotting on slopes for quadrupedal robots,'' in
  \emph{2015 IEEE/RSJ International Conference on Intelligent Robots and
  Systems (IROS)}, 2015, pp. 5129--5135.

\bibitem{Kim_2019_HighlyDynamic}
D.~Kim, J.~D. Carlo, B.~Katz, G.~Bledt, and S.~Kim, ``Highly dynamic quadruped
  locomotion via whole-body impulse control and model predictive control,'' in
  \emph{arXiv preprint arXiv:1909.06586v1)}, 2019.

\bibitem{Hong_2020_RealTime}
S.~Hong, J.-H. Kim, and H.-W. Park, ``Real-time constrained nonlinear model
  predictive control on so(3) for dynamic legged locomotion,'' in \emph{2020
  IEEE/RSJ International Conference on Intelligent Robots and Systems (IROS)},
  2020, pp. 3982--3989.

\bibitem{Ding_2021_RepresentationFree}
Y.~Ding, A.~Pandala, C.~Li, Y.-H. Shin, and H.-W. Park, ``Representation-free
  model predictive control for dynamic motions in quadrupeds,'' \emph{IEEE
  Transactions on Robotics}, vol.~37, no.~4, pp. 1154--1171, 2021.

\bibitem{Hartley_2019_IEKF}
R.~Hartley, M.~Ghaffari, R.~M. Eustice, and J.~W. Grizzle, ``Contact-aided
  invariant extended kalman filtering for robot state estimation,'' \emph{The
  International Journal of Robotics Research}, vol.~39, no.~4, pp. 402--430,
  2020.

\bibitem{Kim_2021_LeggedRobot}
J.-H. Kim, S.~Hong, G.~Ji, S.~Jeon, J.~Hwangbo, J.-H. Oh, and H.-W. Park,
  ``Legged robot state estimation with dynamic contact event information,''
  \emph{IEEE Robotics and Automation Letters}, vol.~6, no.~4, pp. 6733--6740,
  2021.

\bibitem{Hwangbo_2016_ProbabilisticFoot}
J.~Hwangbo, C.~D. Bellicoso, P.~Fankhauser, and M.~Hutter, ``Probabilistic foot
  contact estimation by fusing information from dynamics and
  differential/forward kinematics,'' in \emph{2016 IEEE/RSJ International
  Conference on Intelligent Robots and Systems (IROS)}, 2016, pp. 3872--3878.

\bibitem{Katz_2019_Minicheetah}
B.~{Katz}, J.~D. {Carlo}, and S.~{Kim}, ``Mini cheetah: A platform for pushing
  the limits of dynamic quadruped control,'' in \emph{2019 IEEE International
  Conference on Robotics and Automation (ICRA)}, 2019, pp. 6295--6301.

\bibitem{Hwangbo_2018_PerContact}
J.~Hwangbo, J.~Lee, and M.~Hutter, ``Per-contact iteration method for solving
  contact dynamics,'' \emph{IEEE Robotics and Automation Letters}, vol.~3,
  no.~2, pp. 895--902, 2018.

\bibitem{Schulman_2017_PPO}
J.~Schulman, F.~Wolski, P.~Dhariwal, A.~Radford, and O.~Klimov, ``Proximal
  policy optimization algorithms,'' \emph{CoRR}, vol. abs/1707.06347, 2017.

\bibitem{Sutton_2018_RL}
R.~S. Sutton and A.~G. Barto, \emph{Reinforcement Learning: An Introduction},
  2nd~ed.\hskip 1em plus 0.5em minus 0.4em\relax The MIT Press, 2018.

\bibitem{Reda_2020_LearningToLocomote}
D.~Reda, T.~Tao, and M.~van~de Panne, ``Learning to locomote: Understanding how
  environment design matters for deep reinforcement learning,'' in
  \emph{Motion, Interaction and Games}, ser. MIG '20.\hskip 1em plus 0.5em
  minus 0.4em\relax New York, NY, USA: Association for Computing Machinery,
  2020.

\bibitem{Peng_2018_SimtoReal}
X.~B. Peng, M.~Andrychowicz, W.~Zaremba, and P.~Abbeel, ``Sim-to-real transfer
  of robotic control with dynamics randomization,'' in \emph{2018 IEEE
  International Conference on Robotics and Automation (ICRA)}, 2018.

\bibitem{Hyun_ImplementationOfTrot}
D.~J. Hyun, J.~Lee, S.~Park, and S.~Kim, ``Implementation of trot-to-gallop
  transition and subsequent gallop on the mit cheetah i,'' \emph{The
  International Journal of Robotics Research}, vol.~35, no.~13, pp. 1627--1650,
  2016.

\end{thebibliography}

\end{document}